\documentclass[twocolumn]{svjour3} 

\smartqed  

\usepackage{graphicx}
\usepackage[numbers]{natbib}
\usepackage{amssymb, amsmath}
\usepackage[usenames, dvipsnames]{color}
\usepackage{textcomp, gensymb}
\usepackage[hidelinks=true]{hyperref}
\usepackage{enumitem}
\usepackage{color}
\usepackage{multirow}
\usepackage{xspace}
\usepackage{graphicx,import}
\usepackage{wrapfig}
\usepackage{caption}
\usepackage{subcaption}
\usepackage{algorithm}
\usepackage[noend]{algpseudocode}
\usepackage{xpatch}
\usepackage{mathptmx} 

\algrenewcommand\alglinenumber[1]{{\sffamily\footnotesize#1}}

\makeatletter
\xpatchcmd{\algorithmic}{\itemsep\z@}{\itemsep=0.5ex}{}{}
\makeatother

\algdef{SE}[SUBALG]{Indent}{EndIndent}{}{\algorithmicend\ }%
\algtext*{Indent}
\algtext*{EndIndent}

\DeclareCaptionFormat{myformat}{#3}
\DeclareGraphicsExtensions{.png,.pdf}
\graphicspath{ {Report_Images/} }

\DeclareMathOperator{\argmax}{argmax}

\newcommand{\bvec}[1]{\mathbf{#1}}
\newcommand{\real}{\mathbb{R}}
\newcommand*{\affaddr}[1]{#1} 
\newcommand*{\affmark}[1][*]{\textsuperscript{#1}}
\newcommand{\mct}{$\mu$CT\xspace}

\def\makeheadbox{\relax}

\begin{document}

\title{Improved Workflow for Unsupervised Multiphase Image Segmentation}

\author{Brendan A. West\affmark[1] \and Taylor S. Hodgdon\affmark[1] \and Matthew D. Parno\affmark[1] \and Arnold J. Song\affmark[1]}
\authorrunning{Brendan A. West \and Taylor S. Hodgdon \and Matthew D. Parno \and Arnold J. Song}

\institute{
	Brendan A. West
             	\email{brendan.a.west@erdc.dren.mil} \\ 
	\and
	Taylor S. Hodgdon
	        		\email{taylor.s.hodgdon@erdc.dren.mil} \\
	\and
	Matthew D. Parno
	        		\email{matthew.d.parno@erdc.dren.mil} \\
	\and
	Arnold J. Song
	        		\email{arnold.j.song@erdc.dren.mil} \\
\affaddr{\affmark[1]U.S. Army Corps of Engineers Cold Regions Research and Engineering Laboratory }\\
}

\date{Received: date / Accepted: date}

\maketitle

\begin{abstract}
Quantitative image analysis often depends on accurate classification of pixels through a segmentation process. However, imaging artifacts such as the partial volume effect and sensor noise complicate the classification process. These effects increase the pixel intensity variance of each constituent class, causing intensities from one class to overlap with another.  This increased variance makes threshold based segmentation methods insufficient due to ambiguous overlap regions in the pixel intensity distributions. The class ambiguity becomes even more complex for systems with more than two constituents, such as unsaturated moist granular media.   In this paper, we propose an image processing workflow that improves segmentation accuracy for multiphase systems. First, the ambiguous transition regions between classes are identified and removed, which allows for global thresholding of single-class regions. Then the transition regions are classified using a distance function, and finally both segmentations are combined into one classified image. This workflow includes three methodologies for identifying transition pixels and we demonstrate on a variety of synthetic images that these approaches are able to accurately separate the ambiguous transition pixels from the single-class regions. For situations with typical amounts of image noise, misclassification errors and area differences calculated between each class of the synthetic images and the resultant segmented images range from 0.69-1.48\% and  0.01-0.74\%, respectively, showing the segmentation accuracy of this approach. We demonstrate that we are able to accurately segment x-ray microtomography images of moist granular media using these computationally efficient methodologies.
\end{abstract}


\section{Introduction}
\label{sec:introduction}

\begin{figure}[h]
	\centering
	\includegraphics[width=0.48\textwidth]{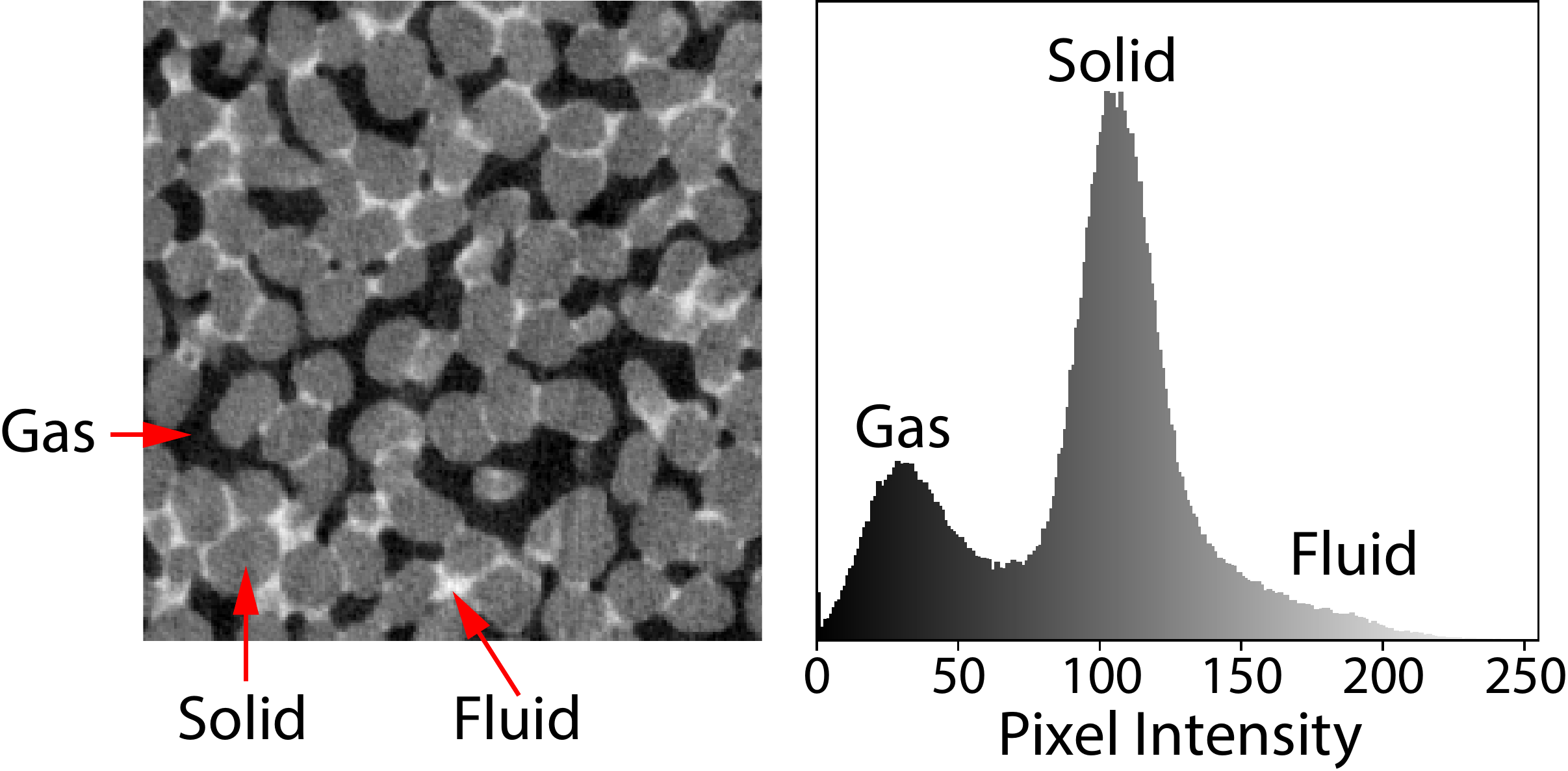}
	\caption{A \mct scan of granular media composed of gas, solid, and fluid phases. The histogram indicates the range and frequency of pixel intensity values within the image.}
	\label{fig:phase_label}
\end{figure}

Given a two-dimensional image, image segmentation is the process of assigning each pixel in the image a particular class, or label. This is an important task in the analysis of experimental observations and is often the first step in much longer scientific workflows \cite{Ketcham2001,Sheppard2004,Kaestner2008,Porter2010}. It is common in many scientific fields to use x-ray microtomography (\mct) to capture micron-scale detail of physical samples. For example, in the study of granular media, segmenting \mct scans of a granular media sample is necessary to study the size and spatial distributions of the sample's components, or to simulate the mechanical behavior of the medium with a numerical model. However, segmentation is difficult for images with more than two constituents, especially when the intensities (i.e., pixel values) of the different components are similar \cite{Sezgin2004,Iassonov2009,Baveye2010}. An example of this occurs in images of moist granular media, where each pixel represents either a gas, solid, or fluid phase, as shown on the left of Figure \ref{fig:phase_label}. The solid and fluid phases attenuate x-rays in a similar manner, and are thus difficult to distinguish in the resultant \mct image. In addition, the fluid phase is less prevalent than the grain phase, which makes the intensity distribution for fluid class confounded with the solid class distribution; in the histogram of Figure \ref{fig:phase_label} it is difficult to even identify the fluid phase. To further exacerbate this problem, raw \mct images are often noisy and typically contain gradual transition zones between classes instead of sharp interfaces. This gradual transition can cause many pixels to be mislabeled in applications with three or more classes.

In images with three components the gradual transitions are particularly problematic for pixels that are in transition between the highest- and lowest-intensity classes. The intensity of these pixels will be midway between the high and low intensities, which is similar to the intensity of the medium intensity class. Figure \ref{fig:phase_label} illustrates the distribution overlap issue in a \mct image of a moist sand sample. The problem areas are those where the gas and fluid phases are adjacent to each other. In these transition regions, pixel intensity does not correspond to a single component, but contains intensity contributions from both the gas and fluid. Therefore, the pixel intensities in the transition between gas and fluid are not distinct from typical intensities of the solid class. This will result in ambiguous class assignment even with segmentation methods that include neighborhood information, e.g., \cite{Vogel1996,Berthod1996,Oh1999,Kulkarni2012}. 

To support more accurate image segmentation in the presence of noise and blurred transitions, we propose a multiphase segmentation workflow that considers both local and global aspects of the image to segment its constituents. Our method uses an initial smoothing step to remove image noise followed by a series of processes to identify those pixels that are in a transition region and those that are not. The transition and non-transition pixels are then segmented separately using the methods described below. The segmented regions are then combined to form a complete segmentation of the image. This workflow is intentionally modular and allows users to interchange different smoothing methods, transition-identification methods, and segmentation algorithms for their particular application. 

We present three methods for identifying transition regions and provide guidance on which method may work best for certain situations. To establish the efficacy of each transition-identification method, we use quantitative measures to compare each method's segmented results for synthetic images with known constituent compositions and levels of Gaussian smoothing and noise. We then apply these methods to a real \mct image of moist sand where the true component at each pixel is unknown.

\section{Background}
\label{sec:background}
Segmenting \mct images of porous media is the subject of several studies \cite{Schluter2010,Andrae2013,Kato2015,Lindquist1999}, and several different methods have been proposed in the literature. Typical image classification methods first use pre-processing to denoise the image before assigning each pixel in the image with one label from a finite set of possibilities (e.g., $\{\text{gas, solid, liquid}\}$). While a comprehensive overview of preprocessing and classification algorithms is not possible here, in part because of the problem-specific nature of many algorithms, below we provide background relevant to our proposed workflow.

\subsection{Edge-Preserving Smoothing Methods}
\label{subsec:smoothing}
Noise is common in \mct images, but can be problematic for segmentation algorithms. We therefore pre-process images with a smoothing step, which helps mitigate the impact of noise. However, naive smoothing filters, e.g., a Gaussian blur filter, can enlarge transition zones, which is also problematic. In the image segmentation context, smoothing methods should reduce noise while also preserving the pixel values along the edges of each component. 

Kaestner and colleagues \cite{Kaestner2008} summarize several smoothing methods used to address noise and transition zones in \mct images of porous media samples. The simplest of these methods are convolution filters, such as the Gaussian blur filter, and spatial filters, such as the mean and median filters. These methods quickly remove noise in image data \cite{Jain1989}, however, they affect all pixels within an image thus failing to preserve the edges, i.e., these filters attenuate the intensity gradient, between components. The authors of \cite{Kaestner2008} also describe a class of edge-preserving filters that use the solutions of diffusive partial differential equations (PDE) as the smoothed image. These approaches have proven useful in many cases \cite{Perona1990,Osher1990,Rudin1992}, but can sometimes require large amounts of memory and can be computationally expensive. While these filters can use up to seven times the memory required to store the initial image, a first-in first-out (FIFO) queue was used by \cite{Kaestner2008} to significantly reduce the amount of required memory. 

One of the most common PDE smoothing methods is the anisotropic diffusion filter (ADF), which uses a diffusion equation with nonlinear diffusivity to reduce noise in an image \cite{Catte1992}.  The PDE takes the form
\begin{equation}
	\frac{\partial u}{\partial t} = \nabla \cdot \left[ g\left(|\nabla_\sigma u |^{2}\right) \nabla u \right],
	\label{eqn:adf}
\end{equation}
where $u$ is the image intensity, $g(\cdot)$ is the nonlinear diffusivity term, and $|\nabla_{\sigma} u|^{2}$ is the squared norm of the image's gradient after being smoothed with a Gaussian filter with standard deviation $\sigma$ \cite{Catte1992,Kaestner2008}. The diffusivity term, $g(\cdot)$, changes based on the magnitude of the image gradient, such that it is small where the gradient is high (such as at edges) and near unity where the gradient is low (such as within homogeneous regions). The result is a filter that smooths within each constituent, but provides minimal smoothing near the interfaces between them.

Another common edge preserving smoothing method is the non-local means algorithm (NLM) proposed by Buades and coworkers \cite{Buades2005}. This method smooths an image by replacing each pixel value with a weighted average of all other pixels in the image. Pixels that have similar Gaussian-smoothed neighborhoods as the target pixel have larger weights than those pixels that do not \cite{Buades2005}.  This smoothing calculation takes the form 
\begin{equation}
	NL[u](x) = \frac{1}{C(x)} \int_{\Omega} e^{-\frac{G_{\sigma}\ast|u(x+.)-u(y+.)|^2(0)}{h^2}} u(y) dy,
	\label{eqn:nlm}
\end{equation}
where $u$ is the image, $x$ is the pixel of interest, $C(x)$ is a normalizing factor, $G_{\sigma}(\cdot)$ is a Gaussian smoothing kernel with standard deviation $\sigma$, $h$ is the filtering parameter, and $y$ corresponds to the other pixels in the image. The $u(x+.)$ and $u(y+.)$ terms correspond to the local neighborhoods around pixels $x$ and $y$, respectively.

The benefit of this smoothing technique is that it takes both global and local information into account as opposed to just local information like many other smoothing filters. NLM smoothing has been used in other studies segmenting porous media \cite{Andrae2013,Schluter2014}. 

\subsection{Segmentation Methods}
\label{subsec:segmentation}
While edge-preserving smoothing methods retain intensity information at the edges, they do not sharpen edges or remove transition zones. Our segmentation approach therefore needs to overcome the transition pixel problem explained earlier. 

Similar to the various smoothing methods in the literature, there are multiple approaches for segmenting the different components of an image. In most cases, this process is challenging due to the gradual transitions between components.  Clustering methods, such as K-Means \cite{Macqueen1967} or Gaussian mixture modeling \cite{Gupta1998} can help identify intensity thresholds that separate components, but are not effective when there is significant overlap between the intensity distributions of each component.  Unfortunately, this overlap is common in \mct images.  As mentioned previously, the small number of fluid pixels creates a class distribution that is essentially indistinguishable from the solid phase distribution. The authors of \cite{Kaestner2008} argue that histogram based segmenting methods are ``unsuitable as a final segmentation method," however we find that these methods, Gaussian mixture modeling in particular, can be useful building blocks for more complex algorithms. 


Several studies have utilized energy based techniques for segmenting images \cite{Hagenmuller2013,Juneja2016}. These methods are based on finding a full segmentation of the image that minimizes an energy function. These methods must consider all pixels in the image to find the minimum-energy segmentation, which is computationally expensive for large three dimensional image sets. While effective in many settings, these types of global segmentation methods are too expensive for our particular purposes. Later we discuss how our segmentation approach uses local deconvolution to incorporate global information in a way that does not couple all of the pixels together, thereby avoiding the cost of common energy-based methods.

Several porous media segmentation studies use morphological operations, such as erosion and dilation, to remove spurious features or fill holes within their segmented results. We avoid these methods because they alter the image data without considering the image's underlying structure.  We have also found that morphological operations can erroneously remove small, but important, regions like the water phases in Figure \ref{fig:phase_label}. 

\subsection{Transition Detection}\label{sec:transdetect}
A significant component of our workflow is identifying transition pixels. Two existing concepts from the literature will prove particularly useful in this regard: gradient-based edge detection (e.g., the Sobel filter), and steerable filters.

Thinking of the image as a function in two variables, the gradient at each pixel is an indication of how rapidly the image intensity is changing at that location. This is commonly used in the image processing community to identify edges, which are simply regions of the image with rapid intensity changes. Many different filters have been developed for approximating the image gradient and detecting edges \cite{Canny1986,Roberts1963,Serra1982,Prewitt1970,Sobel1990}. Of these many options, we employ the well-studied morphological gradient \cite{Serra1982,Maragos1987}, which is fast to compute and available in many image processing packages. While the true gradient is a vector-valued quantity, the morphological gradient approximates the magnitude of the gradient at a pixel as the difference between the maximum intensity and minimum intensity over a small neighborhood around the pixel.

It is well-known that an image gradient is sensitive to noise in the image. However, we have found that an initial edge-preserving smoothing of the image (e.g., ADF or NLM) helps reduce the noise-induced gradient peaks.

The Sobel filter \cite{Sobel1990}, which is a common method for computing the gradient at each point in an image, involves the convolution of an image with two $3 \times 3$ filter images, denoted here by $G_{0^\circ}$ and $G_{90^\circ}$.   These filter images approximate the horizontal (i.e., $0^\circ$) and vertical (i.e., $90^\circ$) components of the gradient.  They are given by
\begin{eqnarray}
G_{0^\circ} & = & \left[ \begin{array}{ccc} 1 & 0 & -1\\ 2 & 0 & 2\\ 1 & 0 & -1 \end{array} \right]\\
G_{90^\circ} & = & \left[ \begin{array}{ccc} 1 & 2 & 1\\ 0 & 0 & 0\\ -1 & -2 & -1 \end{array} \right],
\end{eqnarray}
and satisfy
\begin{equation}
\nabla u(\bvec{x}) = \left[ G_{0^\circ}\ast u(\bvec{x}), \,\, G_{90^\circ} \ast u(\bvec{x})\right].
\end{equation}
Notice that from $\nabla u(\bvec{x})$, we can compute any directional derivative.  Let $G_{\theta}$ denote the filter that would give us the directional derivative in the direction $\theta$.  Thus, 
\begin{equation}
G_{\theta} \ast u(\bvec{x})  = \cos(\theta) \left[G_{0^\circ}\ast u(\bvec{x})\right] + \sin(\theta) \left[G_{90^\circ} \ast u(\bvec{x})\right].
\end{equation}
In this sense, the filters $G_{0^\circ}$ and $G_{90^\circ}$ form a basis for any directional derivative filter. This is a specific example of a steerable filter.   In general, a family of filters is said to be steerable if the filter coefficients at any angle $\theta$ can be represented through a finite number of basis filters
\begin{equation}
G_{\theta} \ast u(\bvec{x}) = \sum_{i=1}^N k_i(\theta) \left[G_{\theta_i} \ast u(\bvec{x}) \right],
\end{equation}
where $N$ is the number of basis filters, $k_i(\theta)$ are weights for each basis filter, and $\theta_i$ are the angles of the basis filters.  Freeman and Adelson \cite{Freeman1991} provide more details on the conditions necessary for steerable filters.  They also show how the dominant orientation can be estimated with quadrature pairs of steerable filters.    As an example of quadrature pairs consider the second derivative of a Gaussian, denoted by $G_{\theta}$, and its Hilbert transform, denoted by $H_{\theta}$.  The energy of the image at any angle is given by
\begin{equation}
E_\theta(\bvec{x}) = \left[ G_\theta \ast u(\bvec{x}) \right] ^2 + \left[ H_\theta \ast u(\bvec{x}) \right]^2.\label{eq:energy}
\end{equation}
To find the orientation of the image at a point $\bvec{x}$, we need to find the angle $\theta$ that maximizes this energy.  Following \cite{Freeman1991}, this expression can be approximated by a truncated Fourier expansion to obtain
\begin{equation}
E_\theta\left(\bvec{x}\right) \approx C_1(\bvec{x}) + C_2(\bvec{x}) \cos(2\theta) + C_3(\bvec{x})\sin(2\theta), \label{eq:energyfourier}
\end{equation}
where the coefficients $C_1(\bvec{x}),C_2(\bvec{x}), C_3(\bvec{x})$ depend on the specific filters $G_{\theta}$ and $H_{\theta}$.  The maximum of \eqref{eq:energyfourier} occurs at an angle $\theta^\ast$ given by
\begin{equation}
\theta^\ast(\bvec{x}) = \frac{1}{2}\tan^{-1}\left(\frac{C_3(\bvec{x})}{C_2(\bvec{x})}\right).
\end{equation}

The phase, denoted by $r_{\theta}(\bvec{x})$, at this angle can also be computed from the quadrature pair coefficients
\begin{equation}
r_{\theta}(\bvec{x}) = \tan^{-1}\left(\frac{G_{\theta} \ast u(\bvec{x})}{H_{\theta} \ast u(\bvec{x})}\right).\label{eq:phase}
\end{equation}
Notice that the phase at angle $\theta^\ast(\bvec{x})$ is an estimate of the distance from the point $\bvec{x}$ to an edge in the image.  This will be used in Section \ref{sec:localdeconv} as part of a local deconvolution scheme.

\section{Algorithmic Components}\label{sec:algcomps}
Here we utilize the background and previous work outlined above to introduce two new concepts needed for our segmentation workflow: non-Gaussian mixture modeling of image intensities and local deconvolution.

\subsection{Non-Gaussian Mixture Modeling}\label{sec:nongauss}
As mentioned above, Gaussian mixture models cannot always effectively model the intensities of pixels in transition zones.   To help overcome this, we extend standard Gaussian mixture modeling approaches with non-Gaussian components that explicitly model intensities of transition pixels.  

A standard Gaussian mixture model has a probability density of the form
\begin{equation}
p_g(u) = \sum_{i=1}^N w_i \phi_i(u), \label{eq:gmm}
\end{equation}
where $N$ is the number of components in the mixture, $\phi_i(u)$ is a normal probability density with mean $\mu_i$ and variance $\sigma_i^2$, and $w_i$ is the probability of component $i$.  Gaussian mixture models are in the exponential family of models and provide an efficient and flexible framework for unsupervised classification.  However, to accurately characterize complex distributions, like those coming from \mct intensity images, a large number of components $N$ are required.  This decreases the interpretability of the resulting model.  For example, if there are three phases in the \mct image and three components $N=3$ in the mixture model, it is clear that each component in the mixture model corresponds to a phase in the image.  However, in our case $N>3$, because of the transition regions, and it becomes much more difficult to match the mixture model component with the image phase.  

To overcome this interpretability challenge, we propose the use of a non-Gaussian mixture model with specially designed components to model the intensities of transition pixels.  The intensity of a pixel in transition lies between the intensity of the two phases it is transitioning between.  This is because the physical volume within one pixel contains more than one phase; each phase only fills part of the volume of the pixel.  Let $y_{ij}\sim U[0,1]$ be a uniform random variable describing the fraction of a transition pixel that contains phase $i$ in a transition between phases $i$ and $j$.   Let $u_i \sim N(\mu_i, \sigma_i^2)$ denote the pixel intensity distribution for phase $i$.  We assume the intensity of a transition pixel, denoted by $u_{ij}$, is then given by 
\begin{equation}
u_{ij} = y_{ij}(\mu_j-\mu_{i}) + \mu_i + \epsilon_{ij}, \label{eq:transitionrv}
\end{equation}
where $\epsilon_{ij} \sim N(0, \sigma_{ij}^2)$ for some variance $\sigma_{ij}^2$.  The density of $u_{ij}$ is then given by
\begin{equation}
\psi_{ij}(u) = \frac{1}{\mu_j - \mu_i}\left[\Phi\left(\frac{u-\mu_2}{\sigma_{ij}}\right) - \Phi\left(\frac{u-\mu_1}{\sigma_{ij}}\right)\right], \label{eq:transdens}
\end{equation}
where $\Phi(\cdot)$ is the standard normal cumulative density function.  Combining these transition densities with the Gaussian mixture in \eqref{eq:gmm} yields the non-Gaussian mixture density 
\begin{equation}
p_n(u) =\sum_{i=1}^{N-1}\sum_{j=i+1}^{N} w_{ij} \psi_{ij}(u) + \sum_{i=1}^N w_{i} \phi_i(u). \label{eq:ngmm}
\end{equation}

The unknown model parameters: $\mu_i$, $\sigma_i^2$, $\sigma_{ij}^2$, $w_i$, and $w_{ij}$ can be computed with the expectation maximization (EM) algorithm \cite{Dempster1977}.   We approximate the expectation step in the EM algorithm by sampling over pixel classes with a Gibbs sampler.  The initial values for the model parameters are taken from a simple K-Means classification of the image.

\subsection{Local Deconvolution}\label{sec:localdeconv}
Let $\tilde{u}(\bvec{x})$ denote the image intensity at location $\bvec{x}\in\real^2$.  We model the image as a blurred and noisy function of an idealized image $u(\bvec{x})$ such that 
\begin{equation}
\tilde{u}(\bvec{x}) = S(\bvec{x}) \ast u(\bvec{x}) + \epsilon(\bvec{x}), \label{eq:ld_obs1}
\end{equation}
where $S(\bvec{x})$ is an isotropic Gaussian kernel and $\epsilon(\bvec{x})$ is a white noise Gaussian process.  From the steerable filter, we can get the orientation at any point as well as the filter coefficients at that orientation.  Mathematically, this gives  
\begin{eqnarray}
\tilde{g}(\bvec{x}) &=& G(\bvec{x}, \theta^\ast(\bvec{x}) ) \ast \left[S(\bvec{x}) \ast u(\bvec{x}) + \epsilon(\bvec{x})\right]\label{eq:ld_obs2}\\
\tilde{h}(\bvec{x}) &=& H(\bvec{x}, \theta^\ast(\bvec{x}) ) \ast \left[S(\bvec{x}) \ast u(\bvec{x}) + \epsilon(\bvec{x}) \right],\label{eq:ld_obs3}
\end{eqnarray}
where $\theta^\ast(\bvec{x}) $ is the orientation at $\bvec{x}$ that maximizes the energy in \eqref{eq:energy}.  Notice that \eqref{eq:ld_obs1}, \eqref{eq:ld_obs2}, and \eqref{eq:ld_obs3} provide three complementary observations of $u(\bvec{x})$.  We will use these information sources to estimate the values of the idealized image $u(\bvec{x})$, but first we will parameterize $u(\bar{\bvec{x}})$ in the local area around $\bvec{x}$.

\begin{figure}[h]
	\centering
	\includegraphics[width=0.35\textwidth]{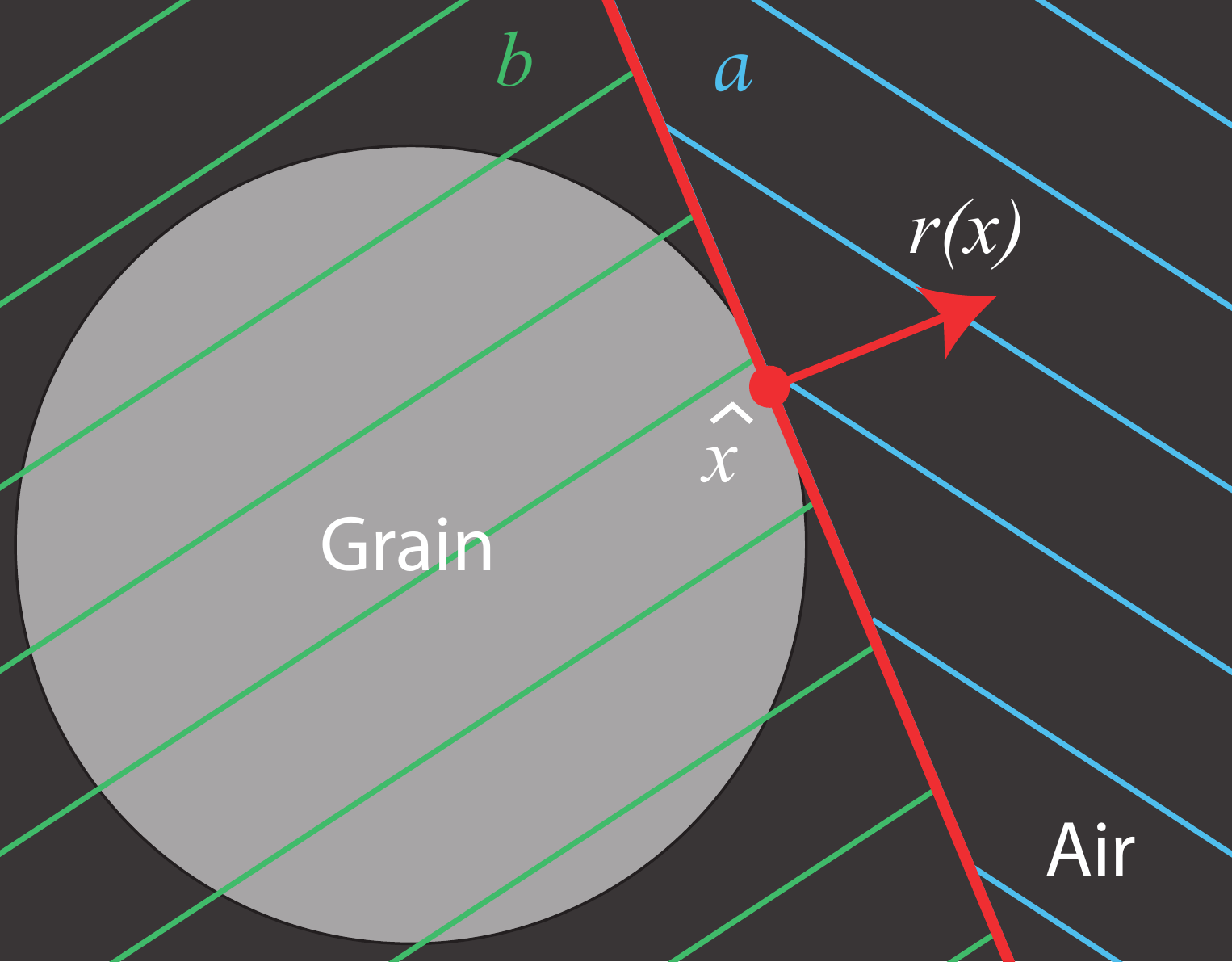}
	\caption{Schematic diagram illustrating the local deconvolution along the contact between grain and gas phases at point $\hat{\bvec{x}}$.  $\bvec{r}(\bvec{x})$ is the vector perpendicular to the edge orientation at point $\hat{\bvec{x}}$, the green hashed area represents where $b(\bvec{x})$ is defined, and the blue hashed area represents where $a(\bvec{x})$ is defined, as in Equation \ref{eq:ideal}.}
	\label{fig:local_dec}
\end{figure}

Near a point $\bar{\bvec{x}}$ in the image, assume that the idealized image $u(\bvec{x})$ is given by as a step function along a particular direction $\bvec{r}(\bvec{x})$.  More specifically, assume that in the neighborhood around $\bvec{x}$ the idealized image is given by
\begin{equation}
u(\bar{\bvec{x}}) \approx \left\{ \begin{array}{cc} a(\bvec{x}) & \bvec{r}(\bvec{x})\cdot \left[\bvec{x} - \hat{\bvec{x}}\right] \geq 0\\ b({\bvec{x}}) & \bvec{r}(\bvec{x})\cdot \left[\bvec{x} - \hat{\bvec{x}}\right] < 0 \end{array}\right. . \label{eq:ideal}
\end{equation}
This local approximation of the idealized image is illustrated in Figure \ref{fig:local_dec}, where $\bvec{r}(\bvec{x})$ is perpendicular to the angle that maximizes the energy and $\hat{\bvec{x}}$ is any point in a transition region.  Note that  $ \bvec{r}(\bvec{x}) \cdot \left[\bvec{x} - \hat{\bvec{x}}\right] $ represents the signed distance between $\bvec{x}$ and the nearest transition; it can therefore be estimated from the phase of the steerable filter.  

 Using the form in \eqref{eq:ideal}, it is possible to compute \eqref{eq:ld_obs1}--\eqref{eq:ld_obs3} analytically and obtain the linear system
\begin{equation}
\left[\begin{array}{c}
\tilde{u}(\bvec{x}) \\
\tilde{g}(\bvec{x})\\
\tilde{h}(\bvec{x})\\
\end{array}\right] = \left[\begin{array}{cc} w_{ua}(r_{\theta^*}) & w_{ub}(r_{\theta^*})\\ w_{ga}(r_{\theta^*}) & w_{gb}(r_{\theta^*}) \\ w_{ha}(r_{\theta^*}) & w_{hb}(r_{\theta^*})\end{array}\right] \left[\begin{array}{c} a(\bvec{x})\\ b(\bvec{x})\end{array}\right]  + \epsilon_{ab},\label{eq:deconv}
\end{equation}
where $r_{\theta^*}\approx \bvec{r}(\bvec{x}) \cdot \left[\bvec{x} - \hat{\bvec{x}}\right] $ is the steerable filter phase from \eqref{eq:phase} computed at point $\bvec{x}$ and  
\begin{equation}
\epsilon_{ab} \sim N(0,\Sigma),
\end{equation}
where $\Sigma$ can be computed by applying the steerable filters to a known synthetic image containing a step function and additive Gaussian noise.  The covariance of the filter coefficients on this synthetic image will be $\Sigma$. The covariance matrix $\Sigma$ captures correlation between $\tilde{u}(\bvec{x})$, $\tilde{g}(\bvec{x})$, and $\tilde{h}(\bvec{x})$ that stems from the appearance of $\epsilon(\bvec{x})$ in all three equations \eqref{eq:ld_obs1}, \eqref{eq:ld_obs2}, \eqref{eq:ld_obs3}.  


\begin{figure}[h]
	\centering
	\includegraphics[width=0.4\textwidth]{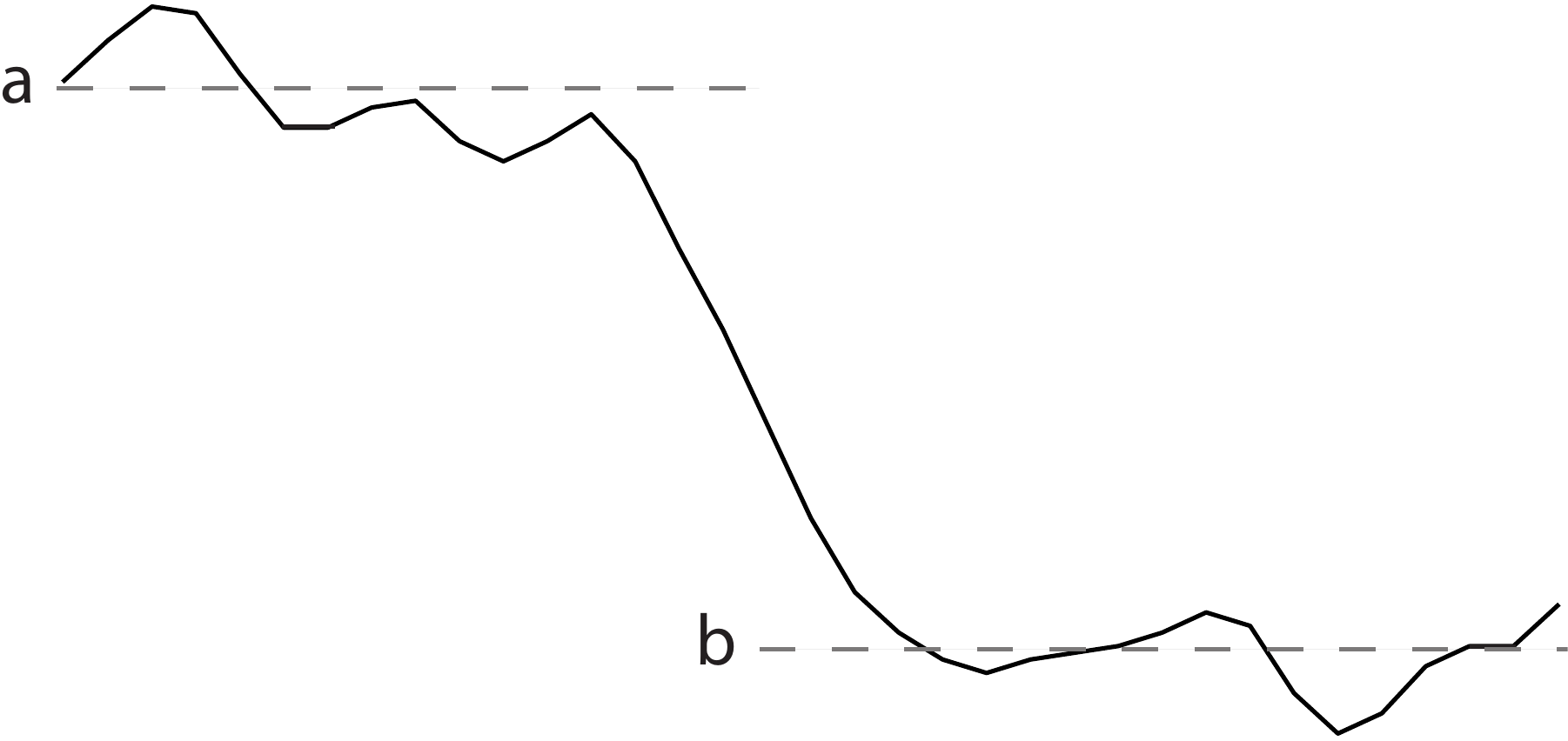}
	\caption{Comparison of idealized image $u(\bvec{x})$ shown by dashed lines at intensities $a$ and $b$ with the true image $u(\bvec{x})$ shown by the solid line.  The intensities shown hear come from the evalution of a two-dimensional image along a line that is perpendicular to an edge.  Our local deconvolution approach provides a method for estimating $a$ and $b$, which can be used to identify transition pixels.}
	\label{fig:ab_decon}
\end{figure}

The expression in \eqref{eq:deconv} defines a weighted least squares problem that can be easily solved to obtain estimates of $a(\bvec{x})$ and $b(\bvec{x})$, and thus the values of the idealized image $u(\bvec{x})$. Figure \ref{fig:ab_decon} shows how the transitions in a real image correspond to the estimates of $a(\bvec{x})$ and $b(\bvec{x})$. The idealized transition consists of a sharp change from one value to the next, whereas the real image transition is gradual and noisy. With the local deconvolution process we attempt to extract the idealized values.  The values of $a(\bvec{x})$ and $b(\bvec{x})$ provide information needed to decide whether a pixel is in a transition zone or not.  If $a(\bvec{x})\approx b(\bvec{x})$, then the pixel is not in a transition zone.

\section{Method Overview}
\label{sec:method}
Our method incorporates the information presented in Sections \ref{sec:background} and \ref{sec:algcomps}, and consists of four main steps: 
\begin{enumerate}
	\item Smooth image to remove noise.
	\item Identify transition pixels and separate image into single-class and transition zone sub-groups. 
	\item Use a Gaussian mixture-method to segment the single-class pixel group.
	\item Segment transition pixels based on minimum distance to a single-class region, as dictated by Euclidean distance. 
\end{enumerate}
The result is a combined segmented image of both the single-class and transition pixels segmented into their respective classifications. 

\subsection{Smooth Image}
\label{subsec:method_smooth}
The first step uses either the ADF or NLM smoothing methods to remove noise within the image while maintaining the intensity values near the component edges. We used a NLM filter from the Scikit Image (Skimage) Python library \cite{Skimage}. Although noise is removed from the regions within each component, the ambiguous transition pixels are left essentially unaltered. This is important because it maintains the original structural information that we use to identify those transition regions. 

\subsection{Identifying Transition Pixels}
\label{subsec:method_transitions}
The next step in the workflow is to identify potential transition pixels. We present three different methods for identifying these pixels based on the techniques outlined in Sections \ref{sec:background} and \ref{sec:algcomps}.

\subsubsection{Local Deconvolution Classification}
\label{subsubsec:deconvolution_steerable}
The first method incorporates the steerable filter ideas from Section \ref{sec:transdetect} with the local deconvolution method introduced in Section \ref{sec:localdeconv} and the non-Gaussian mixture modeling described in Section \ref{sec:nongauss}.  

After smoothing, we construct a non-Gaussian mixture model of the image intensity distribution.  The Gaussian (i.e., in-phase) components of the mixture model allow us to identify which phase is most likely for a particular intensity value.  Once the non-Gaussian mixture model is constructed, we apply the local deconvolution scheme to estimate the $a(\bvec{x})$ and $b(\bvec{x})$ values  in \eqref{eq:ideal} for each pixel in the image.   Combined with the mixture model, the values $a(\bvec{x})$ and $b(\bvec{x})$ allow us to classify each pixel into different transition types (i.e. gas-to-grain, grain-to-fluid, or gas-to-fluid). For example, consider the case where a certain pixel has an $a(\bvec{x})$ value that is most likely in the gas class and a $b(\bvec{x})$ value that is most likely from the grain class.  This pixel is most likely an interface between gas and grain components.  Similarly, if $a(\bvec{x})$ and $b(\bvec{x})$ fall within the same class, the pixel is not likely in a transition zone.  
Notice that this approach does not require setting any thresholds or other critical parameters.

\begin{algorithm}
\caption{Local deconvolution classification algorithm for identifying transition pixels.  }
\label{alg:localclass}
\begin{algorithmic}
\Require Gaussian components $w_j\phi_j(\cdot)$ from the non-Gaussian mixture model in \eqref{eq:ngmm}.
\Require Values $a(\bvec{x})$ and $b(\bvec{x})$ from idealized function in \eqref{eq:ideal} used in local deconvolution.
   \For{Each pixel $i$}
   \State $c_a \gets \underset{j}{\argmax} \left[w_j\phi_j\left( a(\bvec{x}_i) \right)\right]$
   \State $c_b \gets \underset{j}{\argmax} \left[w_j\phi_j\left( b(\bvec{x}_i) \right)\right]$
   \If{$c_a \neq c_b$}
   \State $T_i \gets$ \texttt{True}
   \Else
   \State $T_i \gets$ \texttt{False}
   \EndIf
   \EndFor
  \State \textbf{return} Transition indicator $T_i$ for each pixel.
 \end{algorithmic}
 \end{algorithm}
 
\begin{figure}[h]
	\centering
	\includegraphics[width=0.48\textwidth]{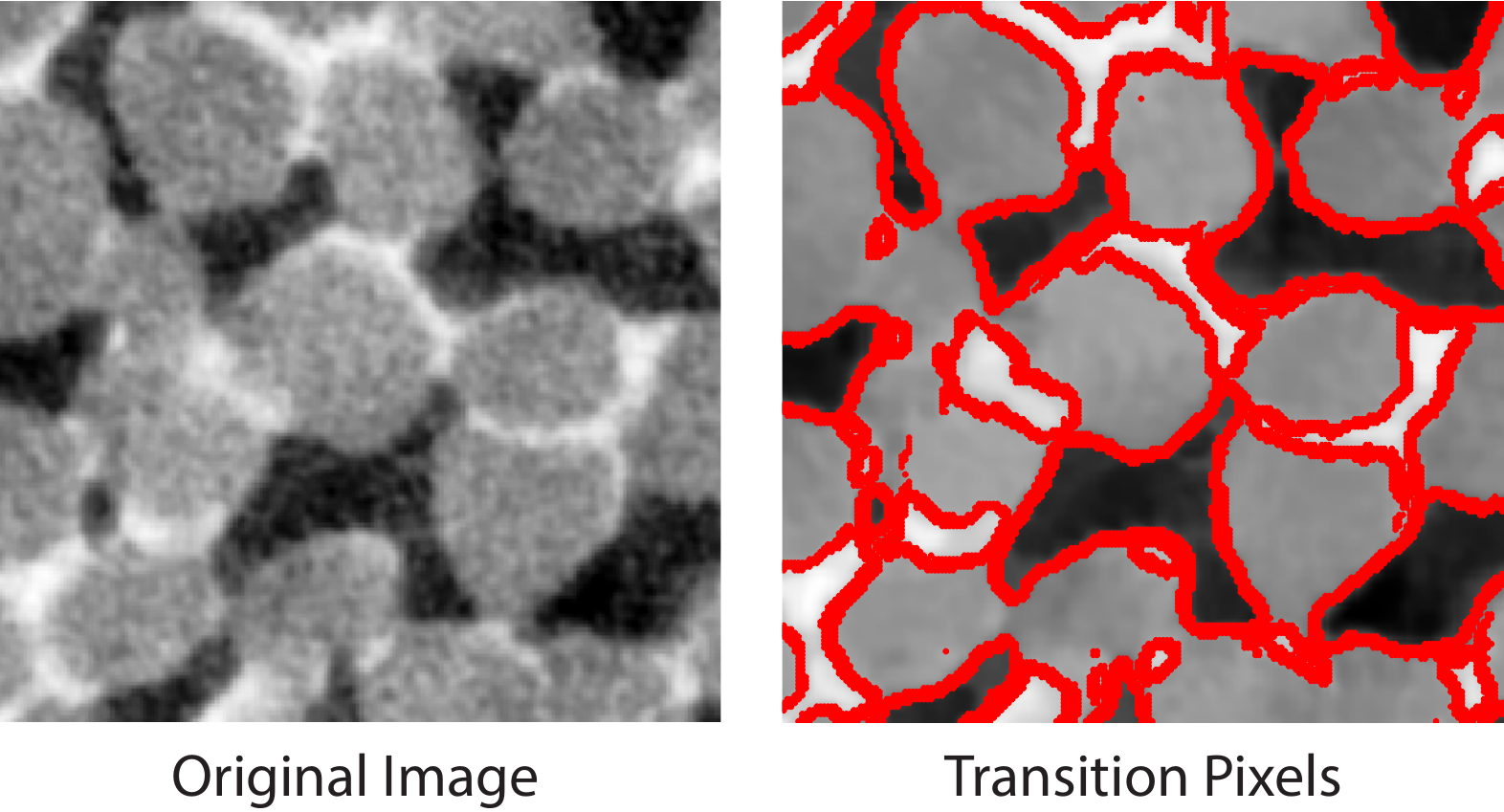}
	\caption{Transition pixels identified using our local deconvolution method. The image underlying the transition pixels is the result of smoothing the original image (left) with NLM.}
	\label{fig:trans}
\end{figure}

Comparing $a(\bvec{x})$ and $b(\bvec{x})$ thus allows us to identify transition regions.   In theory, the values of $a(\bvec{x})$ and $b(\bvec{x})$ could also be used to identity which type of transition region a pixel is in (e.g., gas-grain), but we found these estimates to be inaccurate and only use this approach to identify transition pixels.  Algorithm \ref{alg:localclass} provides a more detailed description of our approach and Figure \ref{fig:trans} illustrates the final transition pixels selected by this method.

\subsubsection{Local Deconvolution Difference}
\label{subsubsec:deconvolution_diff}

Like the above classification approach, this method uses the steerable filter results and local deconvolution scheme to calculate $a(\bvec{x})$ and $b(\bvec{x})$ for each pixel. However, instead of looking at the non-Gaussian mixture model to classify the pixels, the absolute difference between $a(\bvec{x})$ and $b(\bvec{x})$ is used.    If $a(\bvec{x})$ and $b(\bvec{x})$ are significantly different, they likely represent two different components and the pixel is likely in a transition zone.  We therefore use the absolute difference between $a(\bvec{x})$ and $b(\bvec{x})$, $|a(\bvec{x})-b(\bvec{x})|$, as an indication of transition pixels.   If $|a(\bvec{x})-b(\bvec{x})|$ is larger than a prescribed threshold, the pixel is in a transition zone.   To determine an appropriate threshold, we plotted a histogram of all $|a(\bvec{x})-b(\bvec{x})|$ values within the \mct image and determined that $|a(\bvec{x})-b(\bvec{x})| > 20$ was representative of transition pixels in our image.

\subsubsection{Gradient Peak Detection}
\label{subsubsec:method_gradient}
Our last approach for transition detection uses the magnitude of the image gradient as an indicator of transition regions.  If the magnitude of the morphological gradient is larger than a predefined threshold, the pixel is marked as a transition pixel. To determine an appropriate threshold value for separating edges from single-class regions, we plotted a histogram of all gradient values within the \mct image and determined that pixels with gradients $> 30$ belonged to a transition zone. 

\subsection{Segment Single-Class Regions}
\label{subsec:segment_phases}
\begin{figure}[h]
	\centering
	\includegraphics[width=0.48\textwidth]{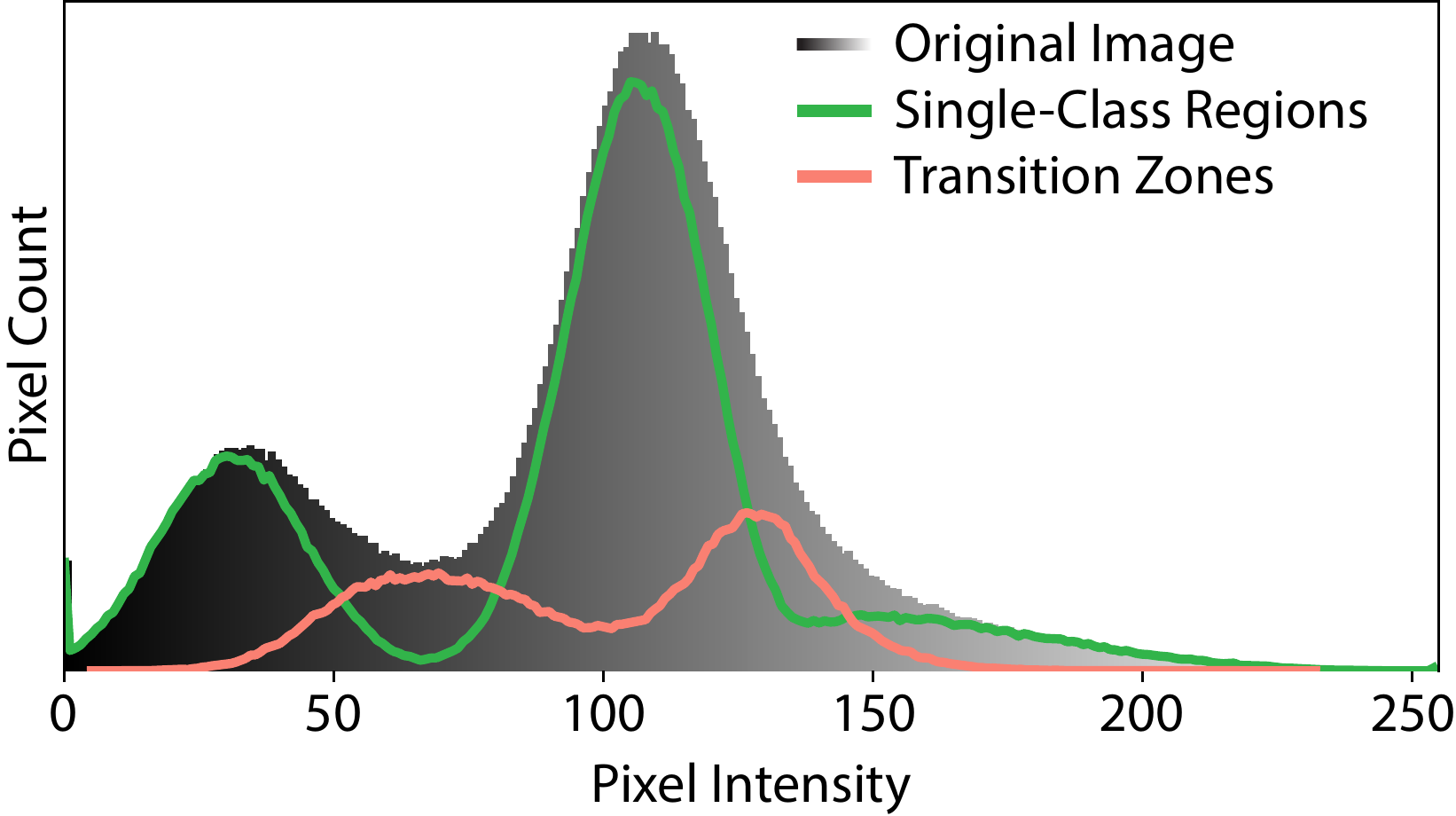}
	\caption{The grayscale distribution reflects the full intensity distribution in the original image.  The green line represents the single-class distribution computed with a Gaussian mixture model and the red line represents the transition zone distribution.}
	\label{fig:intensity_distribution}
\end{figure}

With the transition zones identified and removed from the smoothed image, segmenting the single-class regions is quite simple because the intensity distribution of the phases are more distinct. This step uses a three-component Gaussian mixture model approach to calculate thresholds between each constituent. Figure \ref{fig:intensity_distribution} shows the distributions of both the single-class regions and transition zones after being identified and separated. After removing transition pixels, the underlying Gaussian-like distributions corresponding to each phase become more apparent.  This includes the fluid class, which is confounded with the tail of the grain distribution in the raw intensity histogram.  The transition pixel intensity distribution is bimodal with each mode corresponding to the transitions between gas-to-grain and grain-to-fluid. What this curve does not show is a third mode corresponding to the gas-to-fluid transition since it spans the two other transition types.

\subsection{Segment Transition Regions}
\label{subsec:segment_transitions}
\begin{figure}[h]
	\centering
	\includegraphics[width=0.48\textwidth]{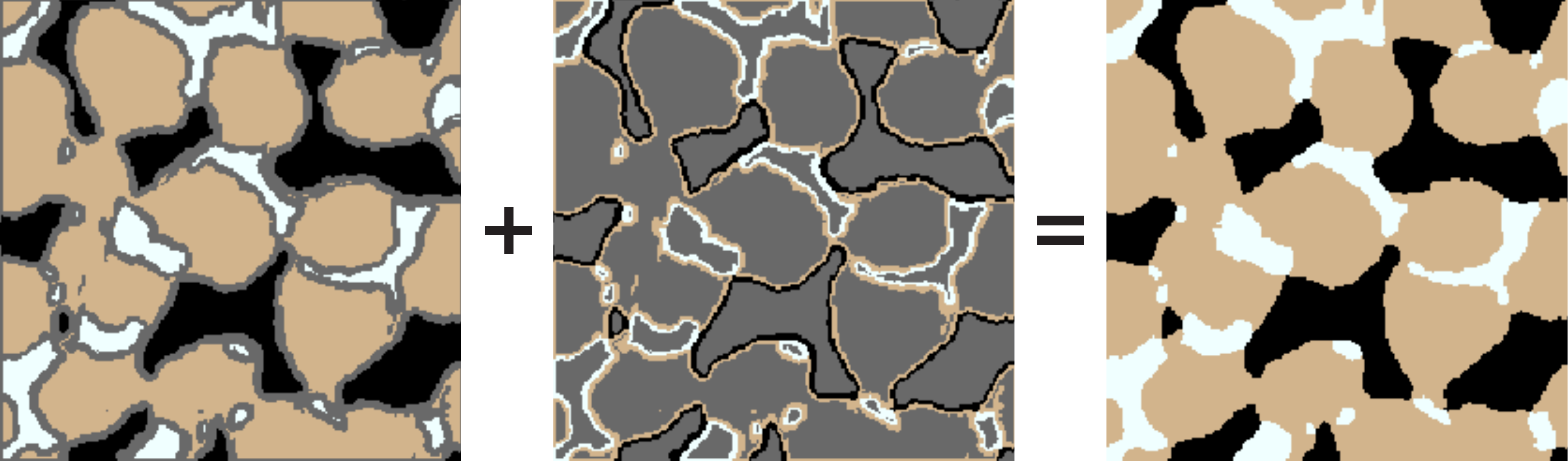}
	\caption{Result of the single-class and transition region segmentations. The final segmented image is the combination of both segmentations. The dark gray regions in the images corresponds to regions with no classification.}
	\label{fig:seg_regions}
\end{figure}

The approaches described in Section \ref{subsec:method_transitions}  are used to identify transition pixels, but not to identify the class of each transition pixel.  To perform this classification, we first compute the distance to the nearest pixel in each single-class region defined in Section \ref{subsec:segment_phases} (e.g., the nearest gas pixel).  The transition pixel is then labeled with the class of whichever single-class is closest. As observed by Hagenmuller \cite{Hagenmuller2013}, the intensity standard deviations of each constituent are the same because it is characteristic of the sensor not the sensed medium. This allows us to assume the constituents spread equally into the transition regions and that it is appropriate to segment the transition pixels based on intensity proximity. Figure \ref{fig:seg_regions} illustrates the results of combining the segmented single-class and segmented transition regions.  In this example, the transition regions were identified with the local deconvolution approach.

\section{Results}

\subsection{Synthetic Comparison of Pixel Identification Methods}
\label{sec:comparison}
\begin{figure}[h]
	\centering
	\includegraphics[width=0.48\textwidth]{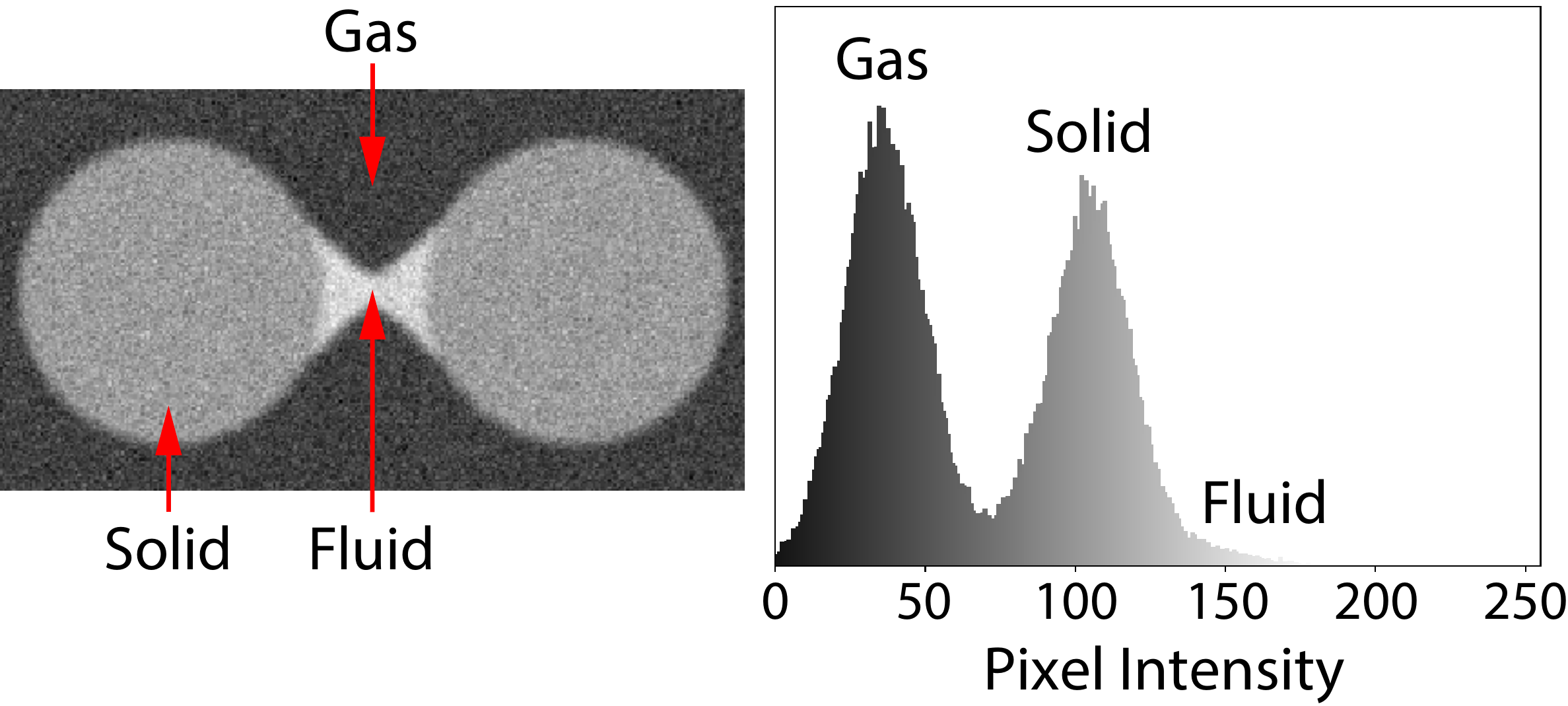}
	\caption{Example of a synthetic test image with corresponding intensity histogram. The Gaussian blur standard deviation and Gaussian noise standard deviation for this image are 3.0 and 13.0, respectively.}
	\label{fig:synthetic_image}
\end{figure}

To evaluate the accuracy of our pixel-identification methodologies we computed several quantitative metrics using a synthetic image representative of granular media \mct images. Artificial test images are useful for calculating comparative metrics because, unlike real \mct imagery, the exact constituent compositions are known. Our base synthetic test image (Figure \ref{fig:synthetic_image}) was composed of two circular grains with a hyperbolic fluid bridge between them, where each phase was assigned the mean intensity value of phases from a \mct scan dataset. 

The synthetic noise and blur artifacts were created by adding Gaussian noise and blurring the image with a Gaussian filter. The segmentation workflow was tested on multiple images with a range of noise levels and transition zone widths to understand how each pixel-identification method responds to each type of image artifact..  The transition widths were controlled by changing the standard deviation of the Gaussian blur filter. We determined from \mct images of moist granular samples that the noise in single-class regions can be approximated with Gaussian noise standard deviations of $\approx14.0$. Similarly, we determined that transition zones can be approximated with Gaussian blur standard deviations of $\approx 2.0$. Therefore, we tested synthetic images with Gaussian noise standard deviations ranging from $10.0$ to $18.0$, and Gaussian blur standard deviations ranging from $1.0$ to $3.0$.

We found that the NLM approach removed noise better than the ADF approach for \mct images of granular media, so we used that smoothing method in all results. After each segmentation, the noisy images were evaluated against the unaltered image in terms of class area fractions and misclassification error (ME). The area fraction provides information regarding the image's total class composition, however it does not verify whether the segmented components are accurate. It is possible to have images with the same area fractions that are vastly different from a pixel to pixel perspective. The ME provides complementary information by comparing each classified pixel in the ``true" image with the corresponding pixel in the final segmented image using the expression
 \begin{equation}
	ME = \frac{1}{n}\sum_{i}I(y_i , \hat{y_i}),
	\label{eqn:ME}
\end{equation}
where $n$ is the number of pixels in the image, $I(\cdot,\cdot)$ is an indicator function that is one when both arguments are the same and zero otherwise, $y_i$ is the initial pixel classification and $\hat{y_i}$ is the pixel class after segmentation.  If the two pixels have the same classification then there is no error, and $I(y_i, \hat{y_i}) = 0$, otherwise $I(y_i, \hat{y_i})=1$.  Classified images that are exactly the same have a ME of $0$, whereas images that are exactly dissimilar have a ME of $1$.

\begin{figure}[h]
	\centering
	\includegraphics[width=0.48\textwidth]{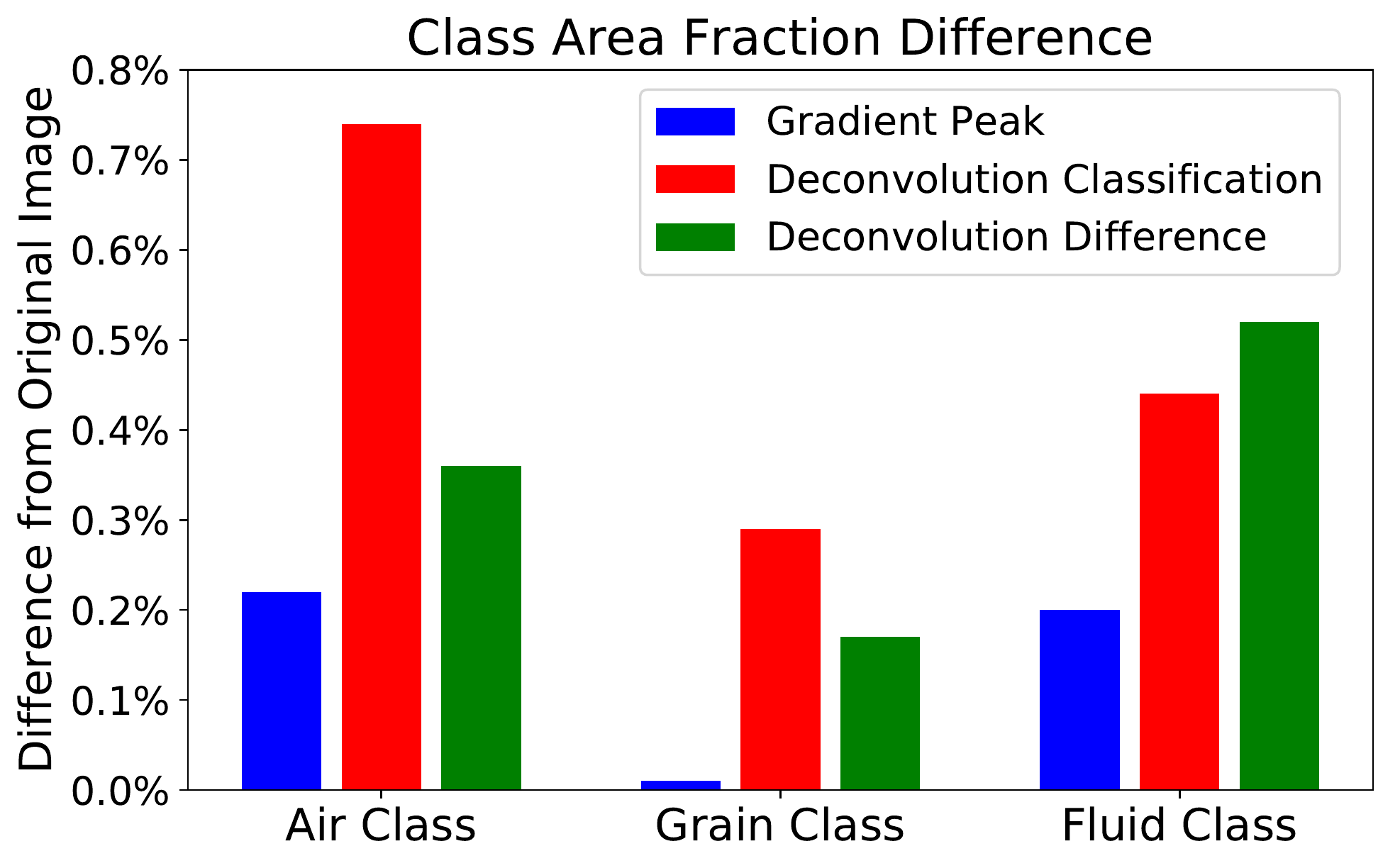}
	\caption{Class area fraction differences from original synthetic image for each transition-identification method. These results are for a synthetic image with Gaussian blur standard deviation of $2.0$ and Gaussian noise standard deviation of $14.0$.}
	\label{fig:Phase_Area_Tests}
\end{figure}

Results from the area fraction calculations are provided in Figure \ref{fig:Phase_Area_Tests}. Overall the three methods provided final segmentations that were close to the true constituent composition of the synthetic images, with the gradient method providing the lowest area differences for all three classes. The deconvolution classification method results in the largest fraction differences in the gas and grain classes of approximately 0.75\% and 0.30\%, respectively. However, the deconvolution difference method produces the highest difference in the fluid class by a small margin. 

\begin{figure}[h]
	\centering
	\includegraphics[width=0.48\textwidth]{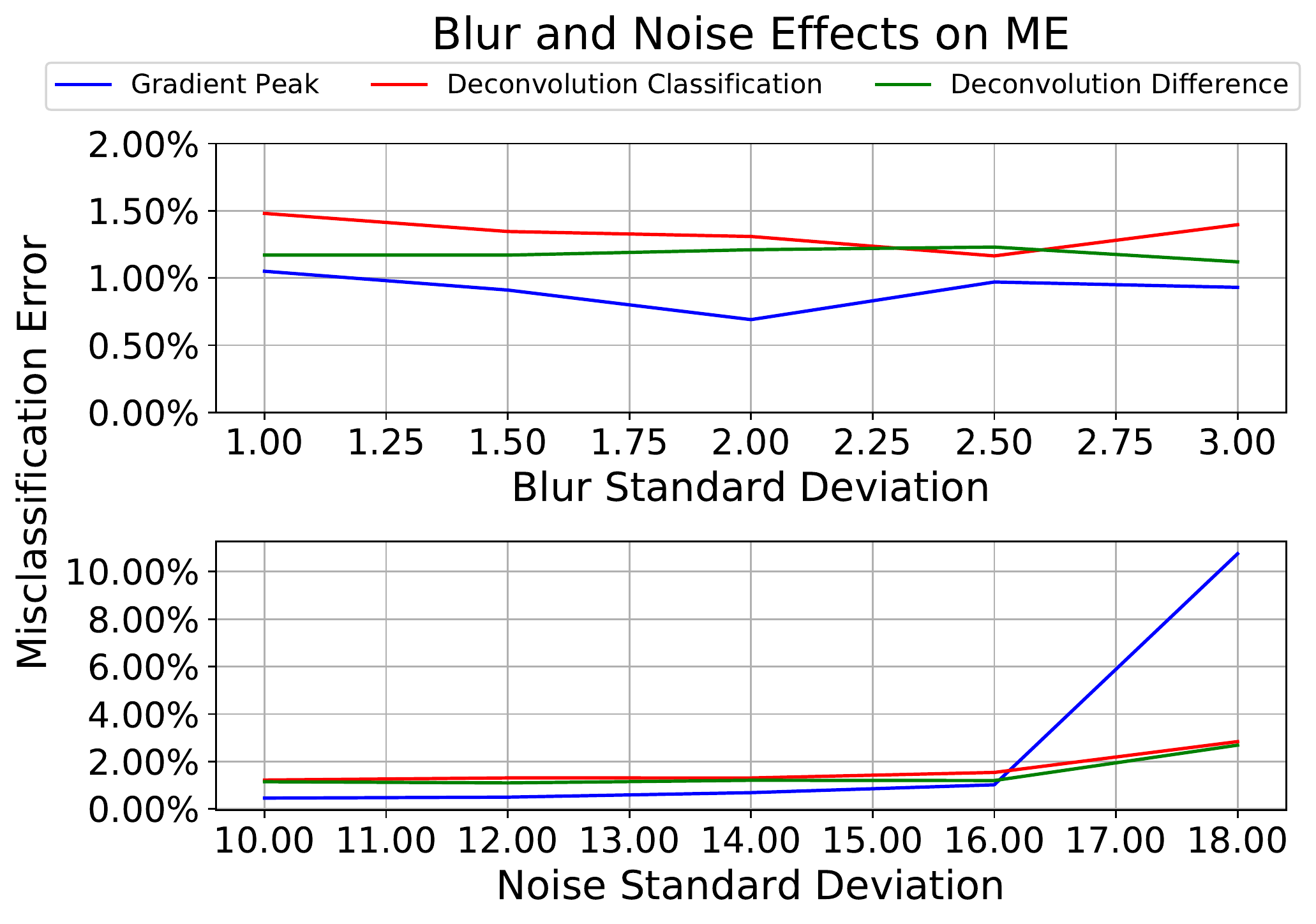}
	\caption{ME results for each method for a range of blur and noise amounts. Top image is for constant Gaussian noise standard deviation of $14.0$ and varying the Gaussian blur standard deviation $1.0$ - $3.0$. Bottom image is for constant Gaussian blur standard deviation of $2.0$ and varying the Gaussian noise standard deviation $10.0$ - $18.0$.}
	\label{fig:ME_Results}
\end{figure}

The ME values appear fairly independent of transition zone width and do not change drastically as the transition zones become larger (top image Figure \ref{fig:ME_Results}). In addition, ME increases slightly as noise increases, however the gradient method ME increases significantly for noise standard deviation of $18.0$ (bottom image Figure \ref{fig:ME_Results}). 

When comparing computational efficiency, the gradient identification method has the shortest average run-time of $1.8$ seconds for a synthetic image that is $160\times 320$ pixels in size. This is significantly faster than the deconvolution difference and classification methods which have average runtimes of $10.3$ and $54.0$ seconds, respectively.  

\subsection{Performance on Real \mct Images}
\begin{figure}[h]
	\centering
	\includegraphics[width=0.48\textwidth]{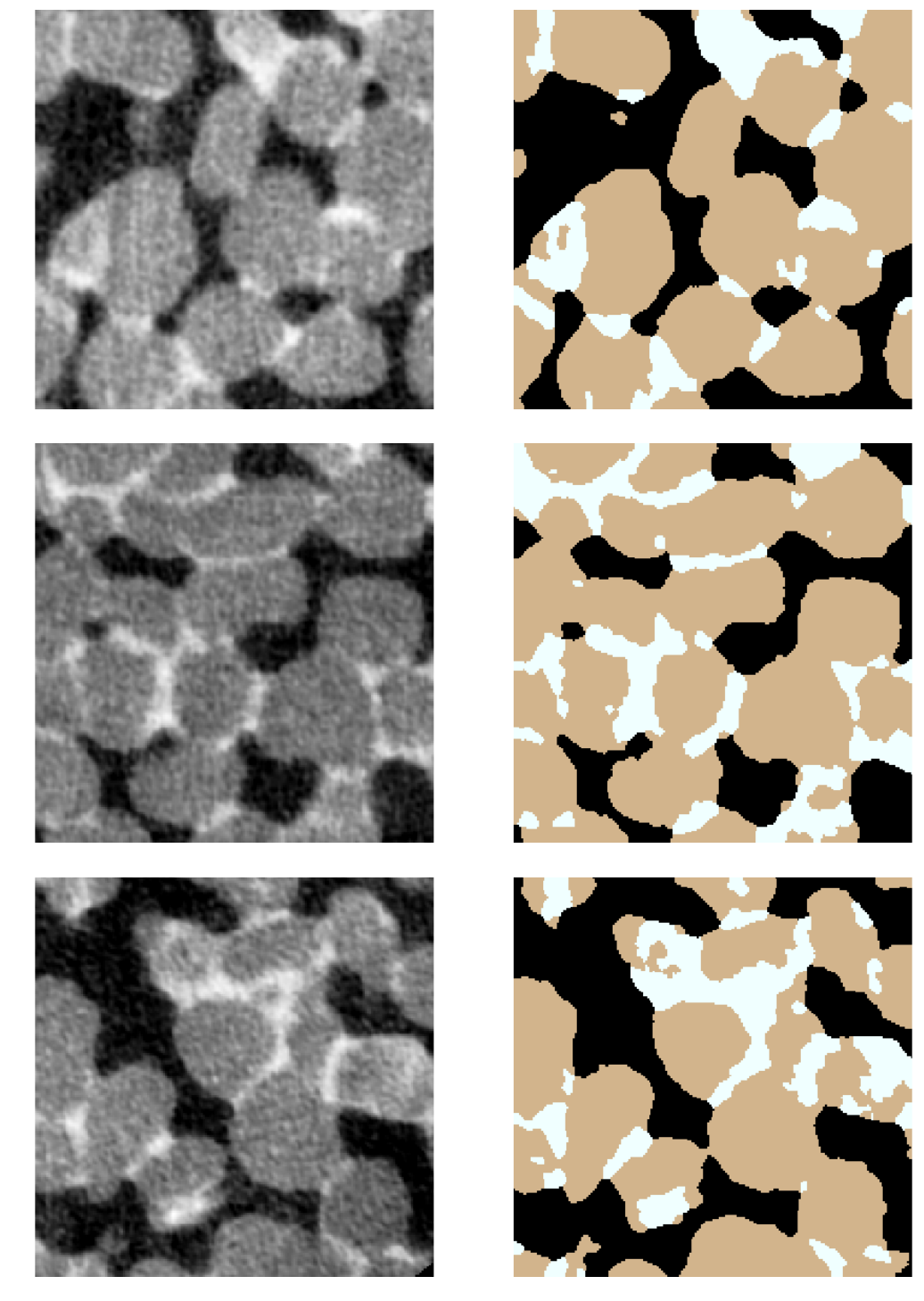}
	\caption{Different sections of a \mct scan and their segmentations using the deconvolution classification method. The black regions correspond to the gas class, the tan regions correspond to the grain class, and the white regions correspond to the fluid class.}
	\label{fig:mct_Examples}
\end{figure}

In addition to the synthetic image tests, we segmented a \mct scan of a moist sand sample. Figure \ref{fig:mct_Examples} illustrates sections of the original \mct scan and their subsequent segmentations. 
This set of segmented images are classified into gas, grain, and fluid classes. With this data we can analyze the composition of the sample and calculate its moisture content. The fluid class of the segmented image makes up $13\%$ of the segmented volume, and the physical sample had a known volumetric fluid content of $12\%$. The difference of $1\%$ is well within measurement error and any human error while preparing the physical sample. This indicates we have achieved good agreement between the physical sample and its segmented counterpart.

\begin{figure}[h]
	\centering
	\includegraphics[width=0.35\textwidth, angle=270]{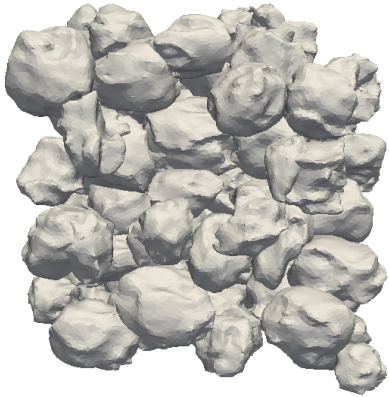}
	\caption{Grain meshes constructed from 300x300x300 section of segmented \mct scan.}
	\label{fig:meshes}
\end{figure}

This segmented dataset will provide the basis for numerical models of the granular samples. As an example, we took a $300\times 300\times 300$ voxel section of the segmented \mct images and created meshes of the different grains using a level set approach (Figure \ref{fig:meshes}). 
Numerical models and computer simulations of real-world phenomena provide researchers with a powerful compliment to theoretical and physical experimentation. In order to provide meaningful results, these numerical simulations require high-fidelity representations of the real-world geometric and material properties that they are modeling. In many cases the simulated properties are merely approximations of the real systems, which is especially true for systems with complex physical shapes and extreme scales. We have shown that we can use our image segmentation methodology to match numerical model geometries with those of actual physical specimens in a precise manner, improving the validation process between numerical and physical experiments. 

\section{Discussion and Conclusions}
\label{sec:discussion_conclusion}
Multiphase image segmentation is challenging due to ambiguity in the transition regions between components, which can lead to pixel misclassification. We have presented a workflow that addresses these transition regions while providing flexibility to substitute different methods for each step of the process. In addition, we have presented the results for three methodologies that identify the ambiguous transition regions within images. 

Of the three methods, the gradient detection method returns the closest segmented class areas to the original image, as well as the lowest ME values for cases with moderate noise. However, it appears much more susceptible to large noise than the other methods, therefore the gradient approach is not recommended when the input images are significantly noisy. The deconvolution classification method results in the highest ME values of the three identification methods and produces the largest area fraction differences for both grain and gas classes. However, this approach had the second closest area fraction of the fluid constituent, which for our images was the most difficult component to segment. The deconvolution difference method provides lower ME values than the deconvolution classification method but not quite as low as the gradient detection method for the gas and grain classes. Of the three methods, the deconvolution difference method seems least affected by both blur and noise in the input images, indicating this method is a good approach for images with significant noise and blurring artifacts. In summary, all methods seem relatively unaffected by increased blur, however they all exhibit decreased accuracy as noise increases. Despite the differences in performance metrics, the area differences are low and the ME values are less than $2\%$ of the entire image for noise levels we would expect to see. This indicates all of these methods provide accurate segmentations of the original image for moderate noise.

While the gradient and deconvolution difference methods performed better than the deconvolution classification approach with respect to area fraction and ME metrics, these methods rely on threshold values that were specifically tuned to produce good results, whereas the classification approach does not rely on such specification. Threshold values that work well for this particular case will likely not work well for others, and the time required to determine adequate thresholds can far outweigh the additional time necessary for the deconvolution classification method to run. The main advantage of the local deconvolution classification approach is that it does not require guidance from the user in order to produce quality image segmentations that can be used for further analyses, such as quantitatively analyzing the image, or initializing numerical simulations.

In summary, we have presented a modular and flexible workflow to segment images composed of multiple constituent phases. In addition, we have presented an unsupervised method for segmenting images that rivals the performance of manually-tailored methods.

\bibliographystyle{spbasic}
\bibliography{Image_Segmentation_Bib}

\begin{thebibliography}{35}
\providecommand{\natexlab}[1]{#1}
\providecommand{\url}[1]{{#1}}
\providecommand{\urlprefix}{URL }
\expandafter\ifx\csname urlstyle\endcsname\relax
  \providecommand{\doi}[1]{DOI~\discretionary{}{}{}#1}\else
  \providecommand{\doi}{DOI~\discretionary{}{}{}\begingroup
  \urlstyle{rm}\Url}\fi
\providecommand{\eprint}[2][]{\url{#2}}

\bibitem[{Andrae et~al(2013)Andrae, Combaret, Dvorkin, Glatt, Han, Kabel,
  Keehm, Krzikalla, Lee, Madonna, Marsh, Mukerji, H.~Saenger, Sain, Saxena,
  Ricker, Wiegmann, and Zhan}]{Andrae2013}
Andrae H, Combaret N, Dvorkin J, Glatt E, Han J, Kabel M, Keehm Y, Krzikalla F,
  Lee M, Madonna C, Marsh M, Mukerji T, H~Saenger E, Sain R, Saxena N, Ricker
  S, Wiegmann A, Zhan X (2013) Digital rock physics benchmarks—part i:
  Imaging and segmentation. Computers \& Geosciences 50:25–32

\bibitem[{Baveye et~al(2010)Baveye, Laba, Otten, Bouckaert, Sterpaio, Goswami,
  Grinev, Houston, Hu, Liu et~al}]{Baveye2010}
Baveye PC, Laba M, Otten W, Bouckaert L, Sterpaio PD, Goswami RR, Grinev D,
  Houston A, Hu Y, Liu J, et~al (2010) Observer-dependent variability of the
  thresholding step in the quantitative analysis of soil images and x-ray
  microtomography data. Geoderma 157(1):51--63

\bibitem[{Berthod et~al(1996)Berthod, Kato, Yu, and Zerubia}]{Berthod1996}
Berthod M, Kato Z, Yu S, Zerubia J (1996) Bayesian image classification using
  markov random fields. Image and vision computing 14(4):285--295

\bibitem[{Buades et~al(2005)Buades, Coll, and Morel}]{Buades2005}
Buades A, Coll B, Morel JM (2005) A non-local algorithm for image denoising.
  In: Proceedings of the 2005 IEEE Computer Society Conference on Computer
  Vision and Pattern Recognition (CVPR'05) - Volume 2 - Volume 02, IEEE
  Computer Society, Washington, DC, USA, CVPR '05, pp 60--65

\bibitem[{Canny(1986)}]{Canny1986}
Canny J (1986) A computational approach to edge detection. IEEE Transactions on
  pattern analysis and machine intelligence PAMI-8(6):679--698

\bibitem[{Catt\'{e} et~al(1992)Catt\'{e}, Lions, Morel, and Coll}]{Catte1992}
Catt\'{e} F, Lions PL, Morel JM, Coll T (1992) Image selective smoothing and
  edge detection by nonlinear diffusion. SIAM Journal on Numerical Analysis
  29(1):182--193

\bibitem[{Dempster et~al(1977)Dempster, Laird, and Rubin}]{Dempster1977}
Dempster AP, Laird NM, Rubin DB (1977) Maximum likelihood from incomplete data
  via the em algorithm. Journal of the royal statistical society Series B
  (methodological) pp 1--38

\bibitem[{Freeman and Adelson(1991)}]{Freeman1991}
Freeman W, Adelson E (1991) The design and use of steerable filters. IEEE
  Transactions on Pattern Analysis and Machine Intelligence 13:891--906

\bibitem[{Gupta and Sortrakul(1998)}]{Gupta1998}
Gupta L, Sortrakul T (1998) A gaussian-mixture-based image segmentation
  algorithm. Pattern Recognition 31(3):315 -- 325

\bibitem[{Hagenmuller et~al(2013)Hagenmuller, Lesaffre, Flin, Calonne, and
  Naaim}]{Hagenmuller2013}
Hagenmuller P, Lesaffre B, Flin F, Calonne N, Naaim M (2013) Energy-based
  binary segmentation of snow microtomographic images. Journal of Glaciology
  59(217)

\bibitem[{Iassonov et~al(2009)Iassonov, Gebrenegus, and Tuller}]{Iassonov2009}
Iassonov P, Gebrenegus T, Tuller M (2009) Segmentation of x-ray computed
  tomography images of porous materials: A crucial step for characterization
  and quantitative analysis of pore structures. Water Resources Research 45(9)

\bibitem[{Jain(1989)}]{Jain1989}
Jain AK (1989) Fundamentals of digital image processing. Prentice-Hall, Inc.

\bibitem[{Juneja and Kashyap(2016)}]{Juneja2016}
Juneja P, Kashyap R (2016) Energy based methods for medical image segmentation.
  International Journal of Computer Applications 146(6)

\bibitem[{Kaestner et~al(2008)Kaestner, Lehmann, and Stampanoni}]{Kaestner2008}
Kaestner A, Lehmann E, Stampanoni M (2008) Imaging and image processing in
  porous media research. Advances in Water Resources 31(9):1174 -- 1187,
  quantitative links between porous media structures and flow behavior across
  scales

\bibitem[{Kato et~al(2015)Kato, Takahashi, Kawasaki, and Kaneko}]{Kato2015}
Kato M, Takahashi M, Kawasaki S, Kaneko K (2015) Segmentation of multi-phase
  x-ray computed tomography images. Environmental Geotechnics 2(2):104--117

\bibitem[{Ketcham and Carlson(2001)}]{Ketcham2001}
Ketcham RA, Carlson WD (2001) Acquisition, optimization and interpretation of
  x-ray computed tomographic imagery: applications to the geosciences.
  Computers \& Geosciences 27(4):381--400

\bibitem[{Kulkarni et~al(2012)Kulkarni, Tuller, Fink, and
  Wildenschild}]{Kulkarni2012}
Kulkarni R, Tuller M, Fink W, Wildenschild D (2012) Three-dimensional
  multiphase segmentation of x-ray ct data of porous materials using a bayesian
  markov random field framework. Vadose Zone Journal 11(1)

\bibitem[{Lindquist and Venkatarangan(1999)}]{Lindquist1999}
Lindquist W, Venkatarangan A (1999) Investigating 3d geometry of porous media
  from high resolution images. Physics and Chemistry of the Earth, Part A:
  Solid Earth and Geodesy 24(7):593 -- 599

\bibitem[{MacQueen(1967)}]{Macqueen1967}
MacQueen J (1967) Some methods for classification and analysis of multivariate
  observations. In: Proceedings of the Fifth Berkeley Symposium on Mathematical
  Statistics and Probability, Volume 1: Statistics, University of California
  Press, Berkeley, Calif., pp 281--297

\bibitem[{Maragos(1987)}]{Maragos1987}
Maragos P (1987) Tutorial on advances in morphological image processing and
  analysis. Optical engineering 26(7):623--632

\bibitem[{Oh and Lindquist(1999)}]{Oh1999}
Oh W, Lindquist B (1999) Image thresholding by indicator kriging. IEEE
  Transactions on Pattern Analysis and Machine Intelligence 21(7):590--602

\bibitem[{Osher and Rudin(1990)}]{Osher1990}
Osher S, Rudin LI (1990) Feature-oriented image enhancement using shock
  filters. SIAM Journal on Numerical Analysis 27(4):919--940

\bibitem[{Perona and Malik(1990)}]{Perona1990}
Perona P, Malik J (1990) Scale-space and edge detection using anisotropic
  diffusion. IEEE Transactions on Pattern Analysis and Machine Intelligence
  12(7):629--639

\bibitem[{Porter and Wildenschild(2010)}]{Porter2010}
Porter ML, Wildenschild D (2010) Image analysis algorithms for estimating
  porous media multiphase flow variables from computed microtomography data: a
  validation study. Computational Geosciences 14(1):15--30

\bibitem[{Prewitt(1970)}]{Prewitt1970}
Prewitt JM (1970) Object enhancement and extraction. Picture processing and
  Psychopictorics 10(1):15--19

\bibitem[{Roberts(1963)}]{Roberts1963}
Roberts LG (1963) Machine perception of three-dimensional solids. PhD thesis,
  Massachusetts Institute of Technology

\bibitem[{Rudin et~al(1992)Rudin, Osher, and Fatemi}]{Rudin1992}
Rudin LI, Osher S, Fatemi E (1992) Nonlinear total variation based noise
  removal algorithms. Physica D: Nonlinear Phenomena 60(1-4):259--268

\bibitem[{Schl{\"u}ter et~al(2010)Schl{\"u}ter, Weller, and
  Vogel}]{Schluter2010}
Schl{\"u}ter S, Weller U, Vogel HJ (2010) Segmentation of x-ray microtomography
  images of soil using gradient masks. Computers \& Geosciences
  36(10):1246--1251

\bibitem[{Schl{\"u}ter et~al(2014)Schl{\"u}ter, Sheppard, Brown, and
  Wildenschild}]{Schluter2014}
Schl{\"u}ter S, Sheppard A, Brown K, Wildenschild D (2014) Image processing of
  multiphase images obtained via x-ray microtomography: a review. Water
  Resources Research 50(4):3615--3639

\bibitem[{Serra(1982)}]{Serra1982}
Serra J (1982) Image analysis and mathematical morphology, v. 1. Academic press

\bibitem[{Sezgin et~al(2004)}]{Sezgin2004}
Sezgin M, et~al (2004) Survey over image thresholding techniques and
  quantitative performance evaluation. Journal of Electronic imaging
  13(1):146--168

\bibitem[{Sheppard et~al(2004)Sheppard, Sok, and Averdunk}]{Sheppard2004}
Sheppard AP, Sok RM, Averdunk H (2004) Techniques for image enhancement and
  segmentation of tomographic images of porous materials. Physica A:
  Statistical mechanics and its applications 339(1):145--151

\bibitem[{Sobel(1990)}]{Sobel1990}
Sobel I (1990) An isotropic 3$\times$ 3 image gradient operator. Machine vision
  for three-dimensional scenes pp 376--379

\bibitem[{Vogel and Kretzschmar(1996)}]{Vogel1996}
Vogel H, Kretzschmar A (1996) Topological characterization of pore space in
  soil—sample preparation and digital image-processing. Geoderma
  73(1-2):23--38

\bibitem[{van~der Walt et~al(2014)van~der Walt, Schönberger, Nunez-Iglesias,
  Boulogne, Warner, Yager, Gouillart, and Yu}]{Skimage}
van~der Walt S, Schönberger JL, Nunez-Iglesias J, Boulogne F, Warner JD, Yager
  N, Gouillart E, Yu Ta (2014) scikit-image: image processing in python. PeerJ
  2:e453

\end{thebibliography}

\end{document}